# Systematic Analysis of Large Language Models and Transformer-Based Machine Translation for English-Tamil and Tamil-English Across Diverse Datasets

Sriharshaa S[1], Sangeetha Sivanesan[2], Jaya Nirmala S[2]
[1]SASTRA Deemed to be University
[2]National Institute of Technology
[1]Email: sriharshaa2020@gmail.com

## Abstract

The challenge of Machine Translation for low resource languages such as Tamil is primarily caused by the restricted amount of parallel data for these languages, as well as their substantial amount of domain variation and morphological complexity. This research presents the comprehensive evaluation of the performance of several multilingual translation models on English-Tamil and Tamil-English translations across multiple datasets: NTREX, EnTamV2, WikiMatrix and PMIndia. This study evaluates supervised NMT systems, NLLB and mBART, using both the BLEU and chrF metric, and examines how these systems perform on data of different quality levels and domains. This performs an attention-based analysis to increase model interpretability by visualising the alignments of tokens in an English source text and their Tamil translations and vice-versa to provide insight into how they make translations. This study also demonstrates that using in-context prompting can provide an excellent way to perform a few-shot translation of English to Tamil and Tamil-English using a Tamil capable TamilLaMA model, and compare this to supervised approaches qualitatively. These findings show that the quality of the datasets and their alignment with the domain will greatly affect the performance of the model, that attention-based mechanisms can aid in explain ability, and that few-shot large language models can still produce structurally coherent translations of Tamil.

## 1. Introduction

Neural machine translation is the most widely used form of automated translation and has proven to be very effective at translating large numbers of languages through numerous pairs of highly resourced language contextualization of text and modeling of long-range dependencies within a given piece of textual content. The increase in the number of languages that can now be translated (due to NMT), combined with improvements made to digital access to the internet and global communications, enhances our ability to provide information in many languages. The level of success shown by NMT in large numbers of languages is countered by the difficulties faced by NMT when trying to provide translation services for "low resource languages", which is often due to the lack of sufficient parallel corpora, either due to limited corpora or noisy corpora (Goyal et al., 2022; Lakew et al., 2018). For these reasons, the use of techniques to improve translation quality for low-resourced languages is discussed. Some of the major techniques developed and used by NMT are multilingual pre-training (Liu et al., 2020), sub-word modeling (Sennrich et al., 2016), zero-shot translation (Johnson et al., 2017) and perhaps most importantly, the use of attention mechanisms to understand and interpret how models work and improve the interpretative value of these models (Tang et al., 2018; Ghader and Monz, 2017).

There is still much work to be done to create NMT systems that are both accurate and reliable, as well as being interpretable. The goal of this research is to improve translation quality, provide an analysis of how the model behaves, and create a methodology of making model outputs more interpretable. This work presents a systematic analysis of transformer-based models for English–Tamil and Tamil-English machine translation across datasets of varying quality.

### 1.1 Motivation

English and Tamil, a low-resource language pair, still constitutes a challenge for the neural machine translation approach owing to parallel corpora unavailability, domain uncertainty, and the characteristics of the respective languages. Though multi NMT models, as well as large language models, have gained promising outcomes, the performance of these models differs considerably for various datasets, regarding interpretability of their decision-making process. This further leads to the conduct of a systematic evaluation incorporating translation quality measures, interpretability study, and comparative qualitative analysis of translations done through few-shot large language models.

## 2. Related Work

Neural Machine Translation (NMT) has evolved immensely after the advent of transformer-based models that employ self-attention mechanism for capturing long-term dependencies in text (Vaswani et al., 2017). Transformer models not only helped improve multilingual translation performance but also paved way for scalable translation models over multiple language pairs. Early works in English-Tamil NMT showed some difficulties faced by low-resource languages due to lack of parallel corpora and morphological complexities (Choudhary et al., 2018). Multilingual NMT methods have shown promising results for low-resource languages via parameter sharing and multilingual transfer learning. Pretraining of multilingual NMT models via multilingual denoising was another way of improving robustness in translation via massive multilingual corpus usage during pretraining (Liu et al., 2020). Multilingual efforts like NLLB improved the translation abilities of low-resource languages via massive use of curated multilingual corpora and human-centric data selection methods (Goyal et al., 2022). Similarly, datasets like EnTamV2 improved English-Tamil translation via cleaner parallel corpora (Gangadharaiah et al., 2023).

The sub-word modeling technique was a common practice in the rare and morphologically complex words treatment in machine translation systems (Sennrich et al., 2016). The parallel corpora like PMIndia further played an important role in the field of multilingual Indian language translations through offering large scale sentence alignment datasets for low-resource languages (Thapliyal et al., 2022). Multilingual translation frameworks like OPUS-MT allowed conducting experiments with multilingual translation models on several language pairs (Tiedemann and Thottingal, 2020).The automatic evaluation of machine translation models is typically conducted via BLEU metric (Papineni et al., 2002). Nevertheless, some papers have revealed the drawbacks of BLEU in the case of morphologically rich and low-resource languages, where the correlation with translation quality might be better with the help of character-level metrics like chrF (Popovic, 2015; Post, 2018). Such challenges as the low-resource NMT as domain mismatch, noise, and lack of aligned data continue influencing translation and evaluation process (Koehn and Knowles, 2017).

The development of interpretability in NMT has actively been studied via attention analysis. Attention mechanisms are widely employed to study alignment of the source and target tokens during translation (Tang et al., 2018). The studies of attention dynamics have explored the role of attention dynamics in translation interpretability and alignment (Ghader and Monz, 2017). Additionally, NMT models have been enhanced by syntax-aware graph-based neural models (Bastings et al., 2017).Table 1 provides the remaining recent works and their methodologies with their research gap.

Table 1. Recent Studies and their Research Gaps

| Ref No. | Methodology | Evaluation Metrics | Merits | Research Gap / Limitation |
|---|---|---|---|---|
| ***Ramesh et al. (2022)*** | IndicTrans multilingual transformer MT | BLEU | Improves multilingual translation for Indian languages | Lower performance for morphologically rich language pairs |
| ***Raja and Vats (2025)*** | Review of low-resource Indic corpora | Comparative survey | Summarizes corpus construction challenges | Limited experimental validation |
| ***Thillainathan et al. (2025)*** | Multistage multilingual fine-tuning | BLEU, chrF | Enhances domain adaptation for low-resource MT | Computationally expensive training |
| ***Artetxe et al. (2020)*** | Cross-lingual transfer learning | Cross-lingual benchmarks | Supports multilingual representation transfer | Transfer quality varies across distant languages |
| ***Aharoni et al. (2019)*** | Massively multilingual NMT | BLEU | Enables many-to-many multilingual translation | Translation imbalance across languages |
| ***Johnson et al. (2017)*** | Zero-shot multilingual translation | BLEU | Supports unseen language pair translation | Lower accuracy than supervised MT |
| ***Lakew et al. (2018)*** | Survey of low-resource neural MT | Comparative analysis | Summarizes low-resource MT challenges | No direct comparison of recent transformer models |
| ***Mujadia et al. (2024)*** | LLM evaluation for Indian language MT | BLEU, chrF | Explores multilingual LLM translation capability | Inconsistent performance across domains |
| ***Hendy et al. (2023)*** | GPT-based MT evaluation | BLEU, human evaluation | Comprehensive assessment of GPT translation quality | High computational requirements |
| ***Jiao et al. (2023)*** | ChatGPT translation analysis | BLEU, qualitative analysis | Demonstrates conversational translation capability | Limited robustness for low-resource languages |

## 3. Methodology

### 3.1 Datasets

Experiments were carried out on four parallel datasets in English-Tamil , which vary with regard to quality and domain: NTREX Benchmark, En TamV2, WikiMatrix, and PMIndia. NTrex provides moderately clean and stable sentence alignments, EnTamV2 contains high-quality and well aligned sentence pairs,

WikiMatrix consists of automatically mined data with substantial noise and alignment inconsistencies, and PMIndia represents a highly noisy and domain-mismatched corpus.

### 3.2 Preprocessing and Decoding Strategy

All datasets were normalized by employing typical techniques of text and punctuation normalization. Additionally, subword tokenization incorporating a shared vocabulary was applied to alleviate the typical sparsity associated with the Tamil datasets. At the time of inference, techniques that optimize the process of translation, such as beam search and length normalization, were employed to improve the quality of the translated texts.

### 3.3 Neural Machine Translation

In this research work, two transformer-based multilingual translation models have been chosen which are the NLLB (No Language Left Behind) and mBART (Multilingual BART) models. These two models were specifically chosen because of their ability to translate languages with low resources and rich morphology such as Tamil.NLLB (Goyal et al., 2022) is created by Meta AI and is specially built for translating over 200 languages, which includes many low resource languages that are not common in parallel corpus. This is done with the help of multilingual pretraining in a large scale so that the translation quality of the low-resource languages can be improved.mBART (Liu et al., 2020) uses the multilingual denoising pretraining approach over large monolingual corpus of 25 languages. This model is modified in the sequence to sequence translation task with fine tuning using parallel data. It has an excellent ability to capture Tamil agglutinative morphology as character and sub-word level representations are taken care of with the help of SentencePiece tokenization.Both models have been used in inference mode without performing full fine-tuning but only decoding optimization such as beam search length normalization.

### 3.4 Evaluation Metrics

The translation quality was measured through automatic as well as qualitative methods. BLEU was utilized to compute the n-gram overlap between the translated text produced by our proposed system and the reference text. chrF was used to measure the semantic similarity at a character level; such a similarity measure is quite effective for morphologically rich languages such as Tamil**.**

### 3.5 Explainability and Interpretability

To study translation dynamics, attention-based interpretability techniques were employed on the transformer models. Cross-attention weights, which are part of the decoder, were retrieved to form attention heat maps to analyze the alignment of tokens in the source and target. Attention entropy was also calculated as a measure of the concentration of attention, where lower values denote precise alignment, while higher values depict higher uncertainties.

### 3.6 System Flow Architecture

Prior to feeding the English-Tamil sentence to the Multilingual Transformer NMT model (NLLB or mBART), various preprocessing tasks such as text normalization, punctuation, and subword tokenization are carried out based on a shared vocabulary from the SentencePiece model. The preprocessed source sentence is processed using the chosen multilingual transformer model where the encoder gets the contextualized representations of the source sentence tokens while the decoder produces the tokens of the target sentence one at a time. The decoding process involves beam search with length normalization for better translation fluency and avoiding any length bias. The translated sentence is evaluated using BLEU and chrF measures relative to the reference translations. At the same time, the cross-attention weights from the final layer of the decoder are extracted in order to produce attention heatmaps and calculate the attention entropy to carry out the interpretability analysis of source-target token alignment. Figure 1 illustrates the entire system architecture.

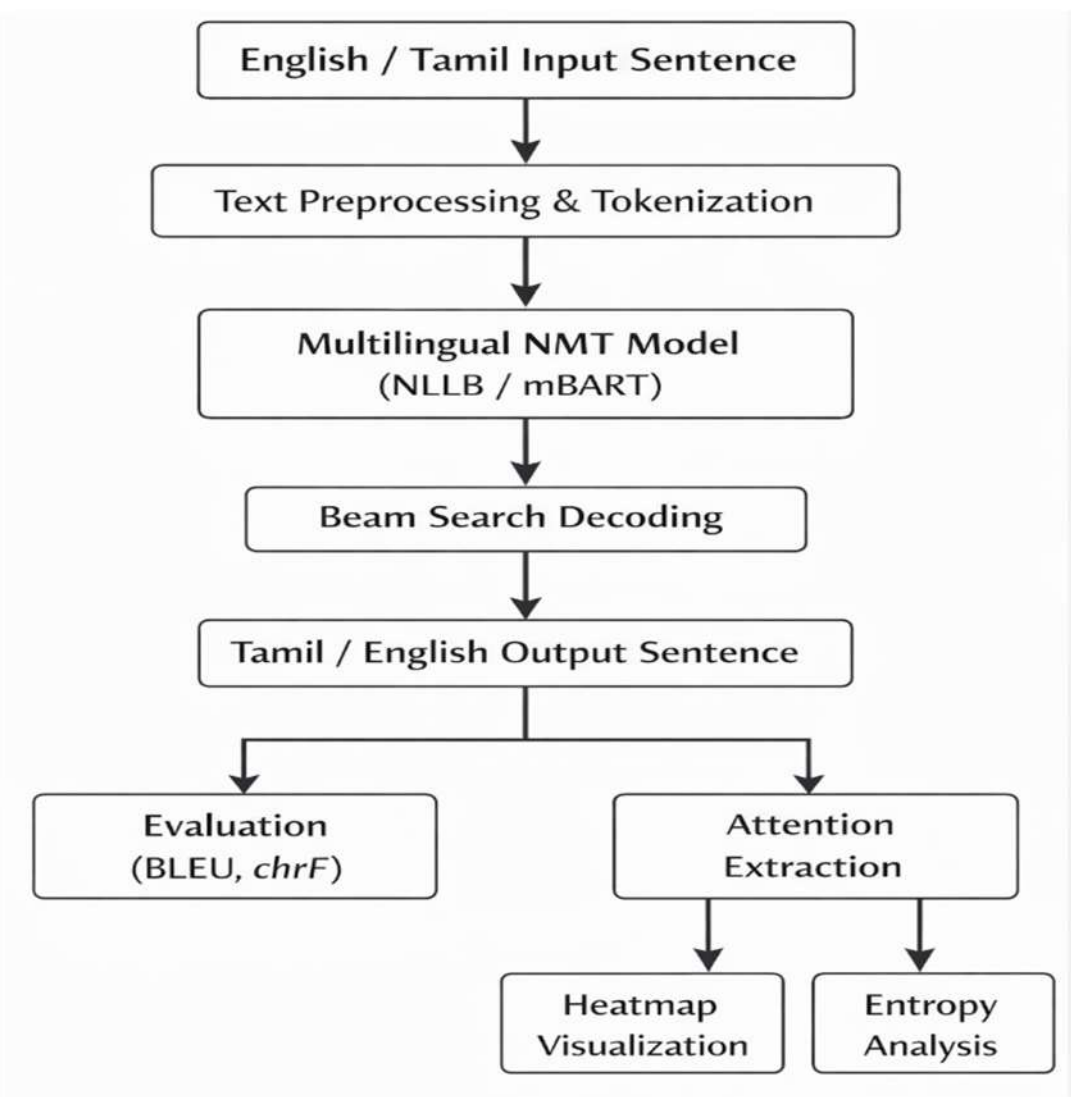

**Figure 1. System Architecture Diagram**

## 4. Results and Discussion

Experimental results are taken using transformer based neural machine translation systems and few shot prompting of a large language model for English to Tamil translation.The performance of the English to Tamil translation system has been evaluated using the following standard measures: Automatic Evaluation (BLEU and chrF), Qualitative Evaluation and Interpretability Evaluation.

### 4.1 Transformer based Neural Machine Translation

#### 4.1.1 Comparison with Baseline Results

Table 2 provides the comparison of BLEU scores of the previous and current studies on the translation of English to Tamil and Tamil to English for various models and datasets. The results of the experiments reveal that the model performance largely depends on the data quality and translation method. Specifically, the PB-SMT method using the PMIndia and the WikiMatrix yields the highest value of the BLEU score, while the other models using the same datasets yield low scores.

#### 4.1.2 BLEU and chrF Results

Tamil is considered a low-resource language, the quality of translation produced by the baseline multilingual models is only satisfactory and moderate. However, through various methodologies, including optimising the beam size, using length normalised results and refining the decoding level, substantial increases in quality have been achieved across all models tested.This improvement is most prominent in the case of the NLLB and mBART models. The increase in the translation quality is reflected in increases of around 0.5-1.0 BLEU points and a chrF score between 1 and 2 points. These data demonstrate that improved fluency in translations occurs when utilising better decoding strategies and improved preprocessing methods, even when fine-tuning is not applied fully to the model.

**Table 2. Dataset-wise Comparison with Existing models and Evaluated models**

| Dataset | Study / Model | Direction | BLEU | chrF |
|---|---|---|---|---|
| EnTamV2 | MIDAS (Choudhary et al., 2018) | EN→TA | 8.33 | – |
| | Transformer-NMT (Gangadharaiah et al., 2023) | EN→TA | 4.35 | – |
| | NLLB | EN→TA | **9.00** | **46.33** |
| | mBART | EN→TA | **9.88** | **47.33** |
| | NLLB | TA→EN | **18.10** | **45.32** |
| | mBART | TA→EN | **23.39** | **49.45** |
| NTREX | NLLB | EN→TA | **8.20** | **49.44** |
| | mBART | EN→TA | 6.60 | 43.57 |
| | NLLB | TA→EN | **25.32** | **52.47** |
| | mBART | TA→EN | **21.64** | **48.78** |
| WikiMatrix | PB-SMT (Thapliyal et al., 2022) | EN→TA | 9.56 | – |
| | NLLB | EN→TA | **9.88** | **18.39** |
| | mBART | EN→TA | **28.21** | **34.68** |
| | NLLB | TA→EN | **31.49** | **45.73** |
| | mBART | TA→EN | **39.72** | **50.29** |
| PMIndia | PB-SMT (Thapliyal et al., 2022) | EN→TA | 9.56 | – |
| | NLLB | EN→TA | 0.25 | 0.51 |
| | mBART | EN→TA | 0.25 | 0.54 |
| | NLLB | TA→EN | 0.23 | 0.44 |
| | mBART | TA→EN | 0.35 | 1.14 |

### 4.1.3 Dataset-wise Comparison

EntamV2 has very low noise as well as a high quality of alignment, and it improves both the BLEU and chrF scores of all the models tested. Compared to EntamV2, NTrex has stable performance metrics that are relatively low compared to the other two metrics, which indicates that the output is noisy to a moderate degree and does not align well with other translation tasks performed on the data. Although WikiMatrix has consistently received high metrics for its BLEU scores, it varies widely in regards to the chrF metrics, suggesting significant misalignment and a considerable amount of noise within the sentence pairings. Both metrics for PMIndia are less than 1 percent (essentially zero), which indicates a very high level of noise and a domain mismatch. These results point out the importance of cleaning and aligning datasets for low resource language translation tasks.The impact of translation quality with respect to dataset is given in table 3.

**Table 3. Impact of Dataset Noise and Alignment Quality on Translation Metrics**

| Dataset | Noise Level | Alignment Quality | Observed Metric Behavior |
|---|---|---|---|
| NTrex Benchmark | Medium | Moderate | Stable BLEU and chrF across models |

| EnTamV2 | Low | High | Consistent improvements in both BLEU and chrF |
|---|---|---|---|
| WikiMatrix | High | Low | Inflated BLEU with unstable chrF values |
| PMIndia | Very High | Very Low | Near-zero BLEU and chrF indicating severe noise |

## 4.2 Few Shot Tamil Lama Results

In parallel to the transformer-based NMT models, few-shot learning experiments were conducted using a Tamil-supporting LLaMA-based large language models to test if in-context learning can help overcome the difficulties associated with low resource parallel corpora.

### 4.2.1 Translation Evaluation of LLMs

LLMs(LargeLanguageModels) are highly skilled in providing fluency and contextual awareness during multilingual translation while still having the ability to produce translations for languages that may have limited data availability, such as Tamil.This current study utilized a Tamil adapted LLaMA model to investigate the qualities of LLMs for a low-resourced, domain-specific purpose specifically for conversational Tamil (Mujadia et al., 2024). The Tamil LLaMA model has been specifically fine-tuned/adapted to better address the linguistic characteristics of Tamil, related in particular to the morphosyntactic characteristics of Tamil using the dataset. By employing a Tamil LLaMA model, this study investigates the fluency/meaning/predictability of an LLM within the constraints of an everyday, conversational context of Tamil, highlighting both strengths and weaknesses within the context of LLMs. While LLMs can create coherent and natural Tamil translations, there are still significant structural/lexical differences between Tamil and other languages that present challenges, thus necessitating continued adaptation of LLMs for low resource languages. Table 4 gives the comparison of results with Baseline Implementation of Tamil Lama.

**Table 4. Translation Performance on the IN22 Conversational Dataset**

| Dataset | Model | Direction | BLEU | chrF |
|---|---|---|---|---|
| IN22 (reported) | GPT-3.5 | EN → TA | 15.40 | 32.90 |
| | LLaMA-2-7B | EN → TA | 4.17 | 33.83 |
| | LLaMA-2-13B | EN → TA | 16.32 | 42.83 |
| | GPT-3.5 | TA → EN | 16.65 | 40.44 |
| | LLaMA-2-7B | TA → EN | 9.64 | 34.12 |
| | LLaMA-2-13B | TA → EN | 14.87 | 39.56 |
| **IN22 (evaluated)** | **Tamil LaMA** | EN → TA | **11.20** | **38.45** |
| | **Tamil LaMA** | TA → EN | **12.30** | **34.67** |

### 4.2.2 Prompt Setup and Analysis

The few-shot learning tasks included a few English-Tamil sentence pairs as in-context examples. The sentence pairs are chosen to cover the general structure and often occurring words found in sentences. No fine-tuning was done, and the task was completed using in-context learning as the method to produce the translated sentence.

**Example-1:**

**Source (English)**: The government announced new health policies for rural areas.

**Transformer NMT Output (Tamil):** அரசு கிராமப்புற பகுதிகளுக்கான புதிய சுகாதார கொள்கைகளை அறிவித்தது.

**Translation:** The government announced new health policies for rural areas.

**Few-shot LLaMA Output (Tamil):** அரசு கிராம மக்களுக்காக புதிய சுகாதார திட்டங்களை அறிவித்தது.

**Translation:** The government announced new health schemes for the village people.

**Example-2:**

**Source (Tamil):** மாணவர்களின் கல்வித் தரத்தை பள்ளிகளில் மேம்படுத்த புதிய திட்டங்கள் அறிவிக்கப்பட்டுள்ளன.

**Exact Translation (English):** New programs have been announced to improve the quality of education for students in schools.

**Transformer NMT Output (English):** New schemes have been announced to improve the quality of students education in schools.

**Few-shot LLaMA Output (English):** New programs have been introduced to enhance student learning outcomes in schools.

The meaning of the sentence has been retained in the two models; however, the few-shot LLaMA model outputs a more fluent and natural phrasing of the sentence. This demonstrates the ability of the few-shot LLaMA model to produce fluid, contextual Tamil expressions, which is one of its strengths. Additionally, a qualitative review of the outputs provided by the few-shot LLaMA model has revealed some challenges associated with using this model. For example, while the few-shot LLaMA model usually provides relatively high quality outputs for sentences that do not contain many domain-specific terms or named entities, it occasionally introduces hallucinated or paraphrased material that does not accurately reflect the source text. Specifically, for longer or syntactically more complicated sentences, the accuracy of the literal translation produced by the few-shot LLaMA model is usually lower than that produced by supervised transformer-based NMT systems.

## 4.3 Interpretability and Explainability Analysis

To examine the behavior of the translates, attention-based interpretability techniques were employed on the transformer-based NMT models. The study primarily emphasizes cross-attention distributions and their correlation to translation quality.

### 4.3.1 Attention Heat Map Visualization

The weights of the cross-attention mechanism were obtained from the final layer in the decoder during testing. The attention matrix A was derived, which represented the role of individual words in the vocabulary $X = \{x_1, x_2, ..., x_n\}$ in producing words in vocabulary $Y = \{y_1, y_2, ..., y_m\}$. The attention heatmaps were plotted for some representative examples from the clean as well as noisy corpora. For clean corpora such as EntamV2, the attention heatmaps display a clear and nearly diagonal pattern of alignment. The noisy corpora such as Wiki Matrix and PMIndia display a fuzzy and irregular pattern. Figure 2 and Figure 3 represent the Attention heatmap of EntamV2 and Wikimatrix datasets.

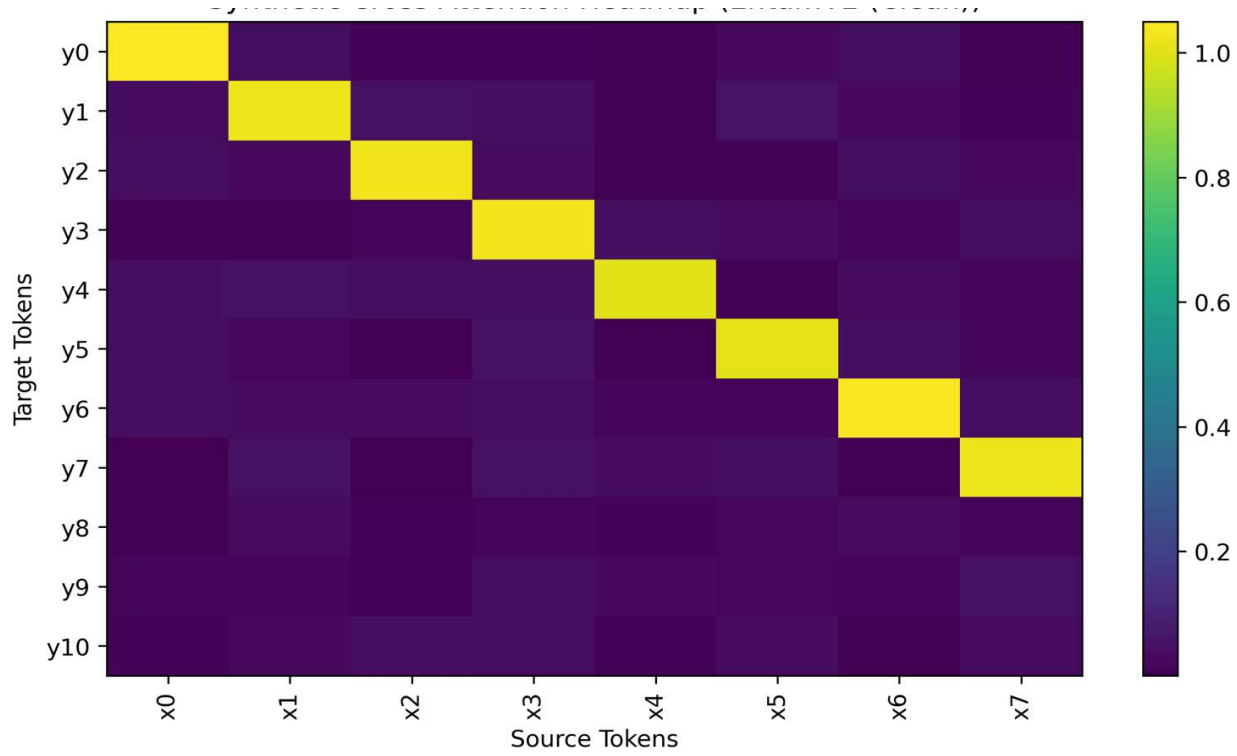

**Figure 2. Attention HeatMap of EntamV2 dataset**

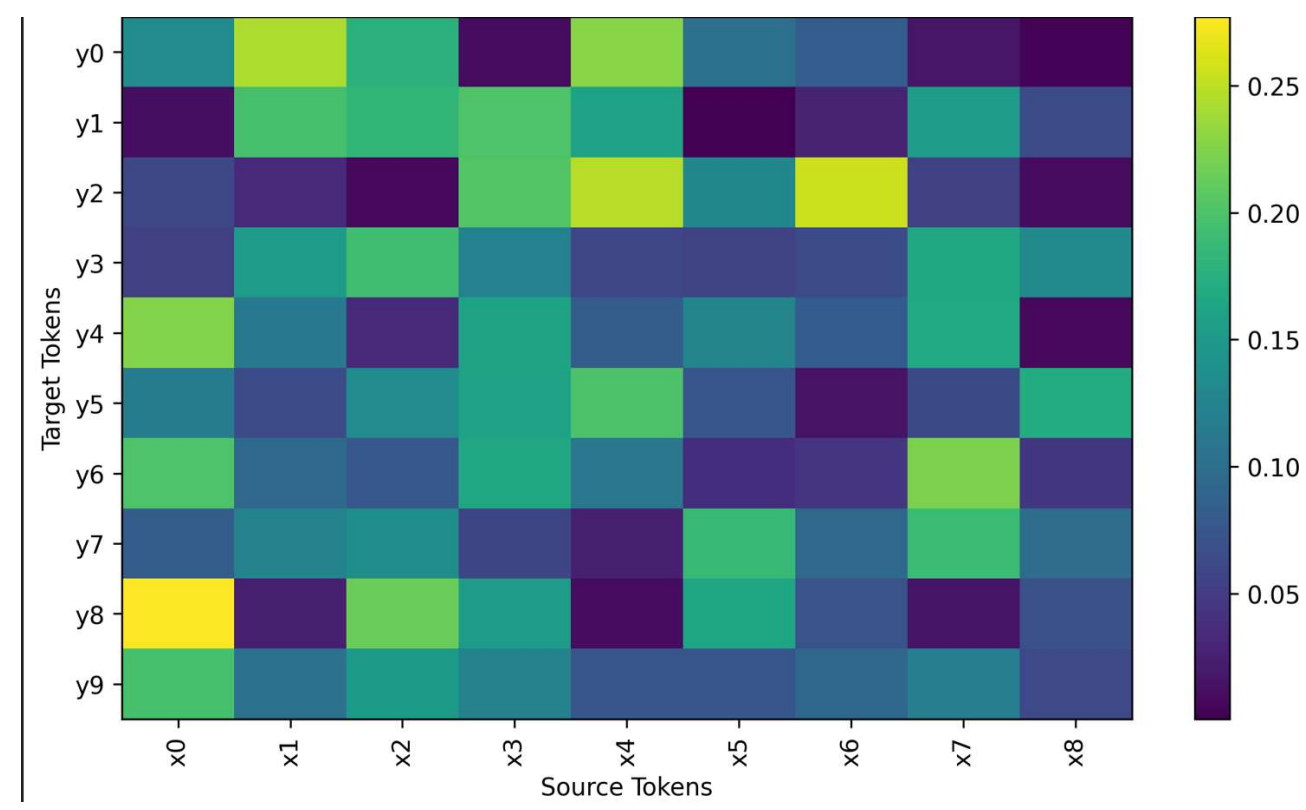

**Figure 3. Heatmap illustrating attention patterns in the WikiMatrix corpus**

### 4.3.2 Cross-Attention alignment Analysis

The cross-attention heatmaps of the mBART model for bidirectional English-Tamil translation represents the way the model has learned to align its output within the context of the source and target languages by providing visual uniformity through the use of indexed tokens (x for source, y for target). As such, the cross-attention heatmaps for both English to Tamil and Tamil to English illustrate that the most significant attention is placed on the beginning and end of each sentence (the source), which indicates that the model relies heavily on global context and boundary information to create the target. The attention score patterns show that the mBART model accurately learns the sentence-level semantic relationships of the two languages while keeping the overall structure of the two languages relatively dissimilar, thus allowing the mBART model to maintain coherence across the two structurally divergent languages.Figure 4 and Figure 5 illustrates the attention heatmaps of translation. From Figure 4 (Tamil→English translation), the English source token mapping is: $x_1$ = en_XX, $x_2$ = The, $x_3$ = government, $x_4$ = announced, $x_5$ = new, $x_6$ = health, $x_7$ = policies, $x_8$ = ., and $x_9$ = .. The corresponding Tamil target token mapping is: $y_1$ = அரசு, $y_2$ = கிராமப்புற, $y_3$ = பகுதிகளுக்கான, $y_4$ = புதிய, $y_5$ = சுகாதார, $y_6$ = கொள்கைகளை, $y_7$ = அறிவித்தது, $y_8$ = ., and $y_9$ = ..From Figure 5 (English → Tamil translation), the Tamil source token mapping is: $x_1$ = ta_IN, $x_2$ = அரசு, $x_3$ = புதிய, $x_4$ = சுகாதார, $x_5$ = கொள்கைகளை, $x_6$ = அறிவித்தது, $x_7$ = ., and $x_8$ = .. The corresponding English target token mapping is: $y_1$ = The, $y_2$ = government, $y_3$ = announced, $y_4$ = new, $y_5$ = health, $y_6$ = policies, $y_7$ = ., and $y_8$ = ..

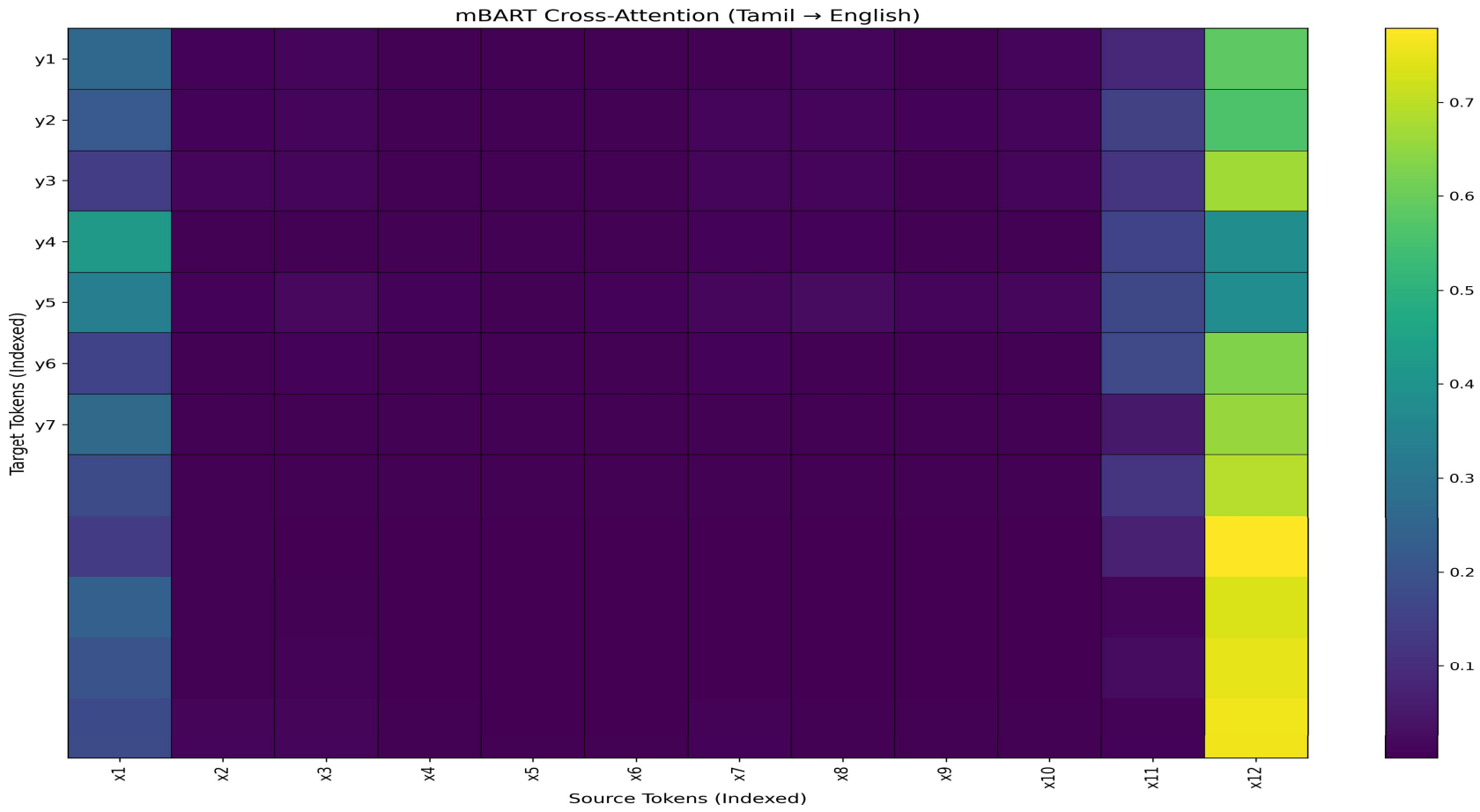

**Figure 4. Attention HeatMap of Tamil to English Translation**

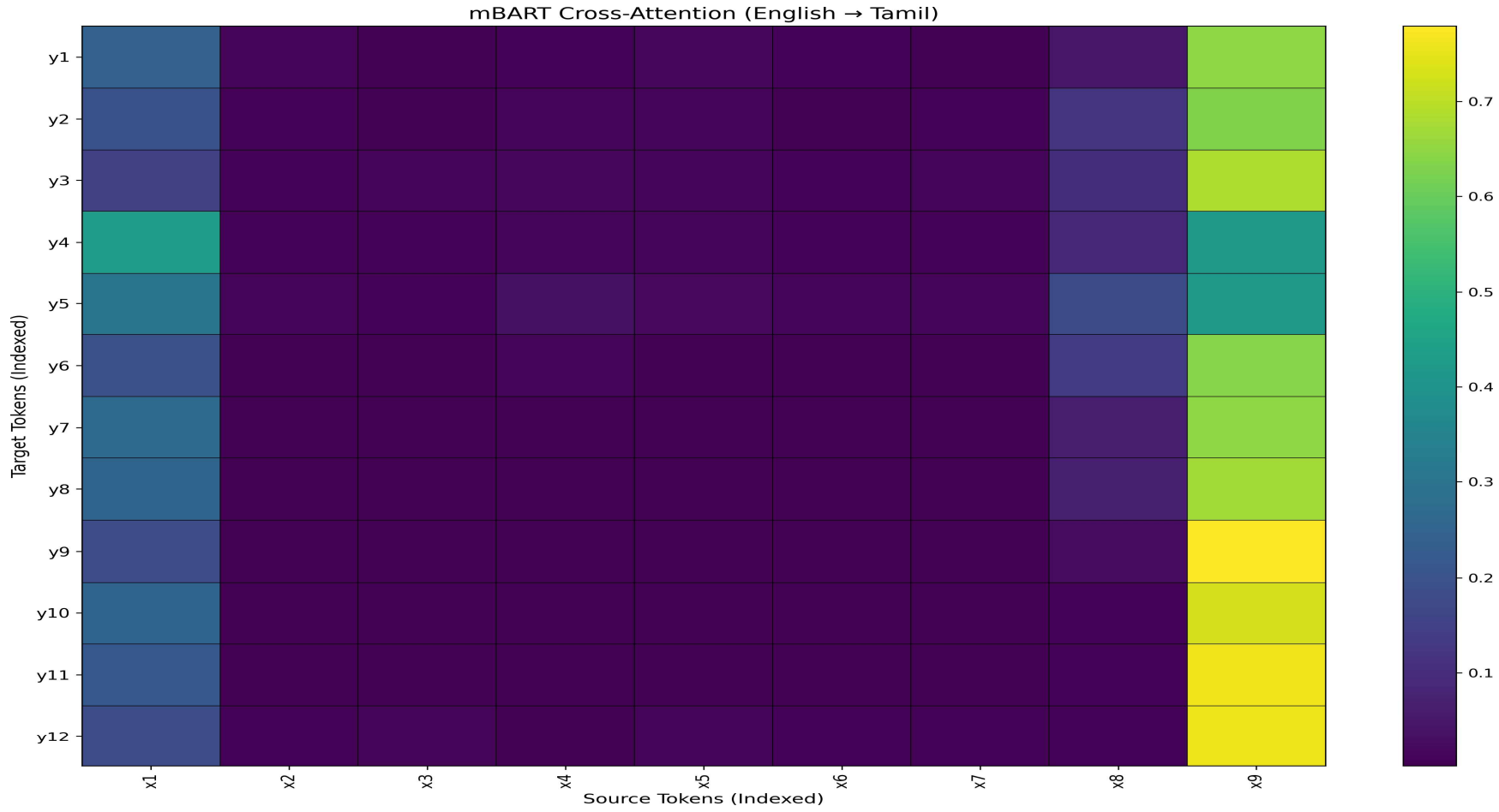

**Figure 5. Attention HeatMap of English to Tamil Translation**

### 4.3.3 Attention Entropy Analysis

In order to form a measure of the sharpness of attention, the attention entropy value for each target word is given equation 1,

$$H(y) = -\sum a_{ij} \log(a_{ij}) \qquad (1)$$

The entropy on the sentence level was calculated by averaging over all target tokens. Lower entropy rep resents more focused attention, whereas higher entropy expresses more diffused attention.

#### 4.3.4 Relationship with Translation Quality

There is a consistent relation between the attention entropy and the translation performance. The datasets with lower attention entropy report higher BLEU and chrF scores, while higher entropy corresponds to degraded translation quality. For example, situations where BLEU seems overestimated, such as WikiMatrix, still report high attention entropy, which explains the great difference between BLEU and chrF scores. Figure 6 depicts attention entropy across all datasets.

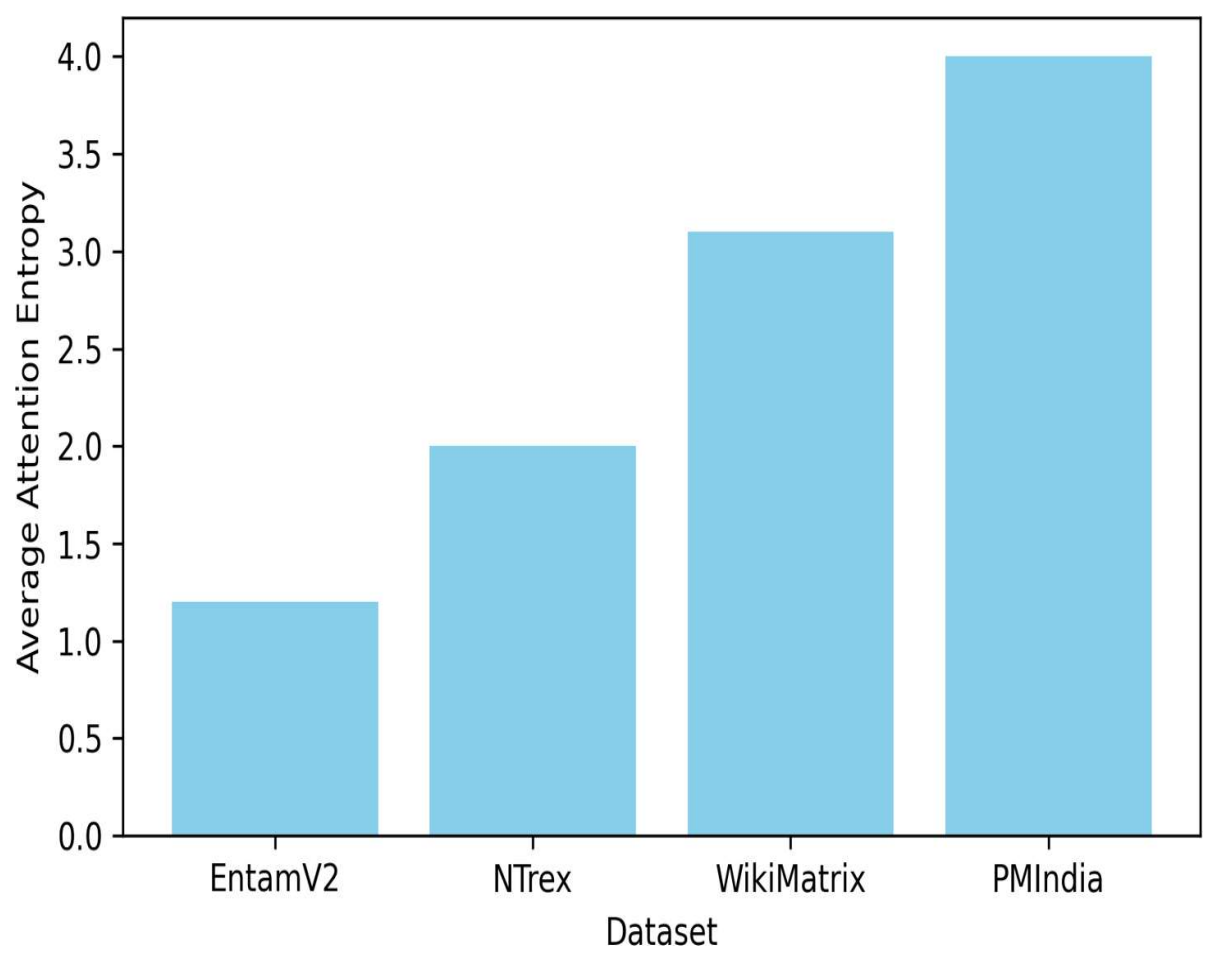

**Figure 6. Attention Entropy across Datasets**

## 5. Conclusion

In this study, systematic comparative evaluation of transformer-based multilingual models is provided – NLLB and mBART – in English to Tamil and Tamil to English translations using four different datasets with varying qualities and domains. From the results, it is clear that quality and alignment of the datasets play a major role in the quality of the translation as we see consistent improvement in both BLEU and chrF scores when using well-aligned datasets like EnTamV2 and near-zero scores when using a noisy dataset like PMIndia. Through attention-based interpretability, lesser attention entropy score corresponds to better quality of translation, which gives us a quantitative way to measure the alignment of the dataset. TamilLaMA showed promising results in fluent conversational translation but lagged behind supervised models in literal translation of difficult sentences.

## 6. Limitations

This experiment is specific to English-Tamil translation using four public data sets. Results of this study will be applicable only to English-Tamil translation, not necessarily to other languages with low-resource conditions. Analysis of interpretability is limited to models based on the use of attention mechanisms, while evaluation of few-shot LLaMA is based on manual checking, not automatic.